\documentclass[10pt,twocolumn,letterpaper]{article}

\usepackage[pagenumbers]{cvpr} 

\definecolor{cvprblue}{rgb}{0.21,0.49,0.74}
\usepackage[pagebackref,breaklinks,colorlinks,allcolors=cvprblue]{hyperref}

\def\confName{CVPR}
\def\confYear{2027}

\title{Verifying Before Enhancing: Target Evidence Chain Verification for Infrared Small Target Detection}

\author{
Jiayi Zuo, Songwei Pei, Shuhuai Wang, Yifei Wu, Shangguang Wang\\
Beijing University of Posts and Telecommunications\\
Beijing, China\\
{\tt\small \{zuojiayi,email2,email3,email4\}@bupt.edu.cn}
}

\begin{document}
\maketitle
\begin{abstract}
Infrared small target detection is essential for early warning, surveillance, remote sensing, and space-based target perception. However, infrared small targets are typically tiny, weakly textured, and boundary-blurred, making them difficult to distinguish from complex backgrounds. Moreover, target-like background structures can produce salient responses across multiple scales, making response strength alone insufficient for reliable target discrimination. 
To address this challenge, we propose VBE-UNet, a Target Evidence Chain Verification Network that assesses candidate reliability beyond response magnitude through complementary evidence cues. 
Specifically, counterfactual background explanation models local target evidence, while joint evidence verification integrates cross-scale support and perturbation-based contribution verification to guide feature recomposition. The resulting reliability estimates guide selective feature refinement, promoting target-relevant responses while reducing unsupported background activations. Extensive experiments on IRSTD-1K, NUDT-SIRST, and NUAA-SIRST demonstrate that VBE-UNet achieves superior detection accuracy and lower false-alarm rates compared with state-of-the-art methods. Ablation studies and evidence reliability analysis further validate the effectiveness of the proposed verification framework.
\end{abstract}    

\section{Introduction}
\label{sec:intro}

Infrared small target detection (IRSTD) plays a critical role in infrared search and tracking systems, with applications in aerial surveillance, maritime monitoring, and remote sensing~\cite{b1,b46,b105}. Unlike visible-light imaging, infrared imaging captures thermal radiation differences, enabling target perception under low illumination and challenging environmental conditions~\cite{b105,b48}. However, infrared small targets typically occupy only a few pixels and exhibit weak intensity, limited structural information, and low signal-to-clutter ratios~\cite{b48,b105}. The absence of distinctive textures and clear boundaries makes it difficult to distinguish genuine targets from surrounding background structures. Moreover, complex backgrounds often contain locally salient patterns that resemble small targets, further increasing the difficulty of reliable detection.

\begin{figure}[t]
    \centering
    \includegraphics[width=\columnwidth]{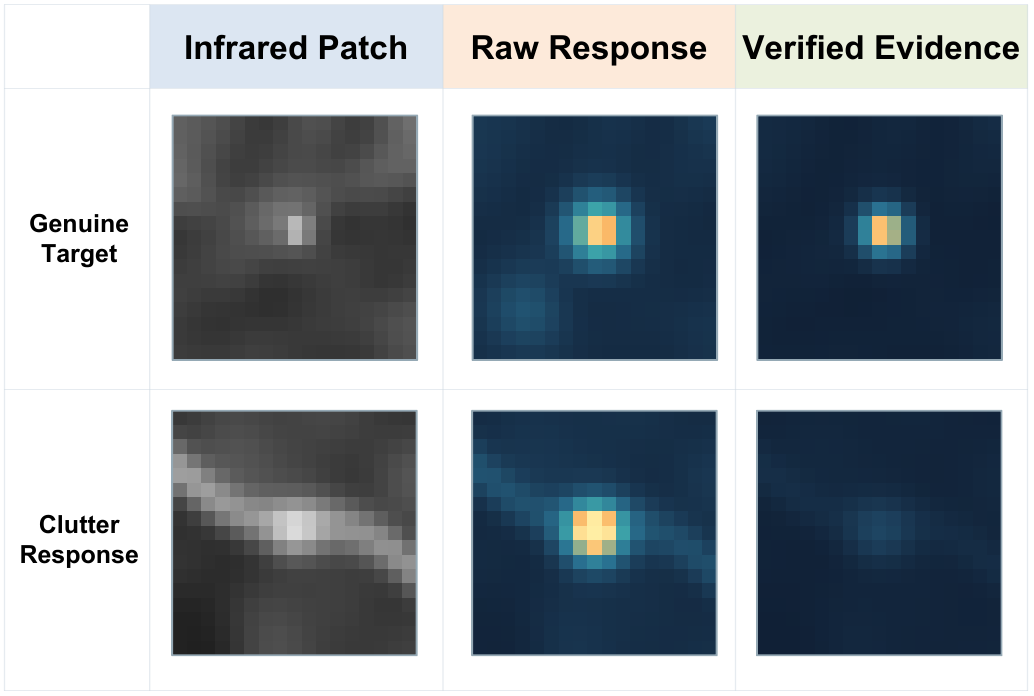}
    \caption{Motivation of target evidence verification. Genuine infrared targets and target-like background structures may exhibit similarly strong responses, but differ in their verified evidence. VBE-UNet leverages these differences to distinguish reliable target responses from background interference.}
    \label{fig_big}
\end{figure}

Existing IRSTD methods have evolved from traditional background filtering, local contrast enhancement, and low-rank decomposition to deep learning-based frameworks~\cite{b52,b53,b20,b55}. Traditional approaches generally rely on handcrafted assumptions about target-background differences, which can be sensitive to complex background variations. In contrast, recent deep models learn discriminative representations through multi-scale feature aggregation, structural priors, and contextual modeling. For instance, DNANet~\cite{b38} and UIU-Net~\cite{b58} employ hierarchical feature fusion to preserve weak target information, while ISNet~\cite{b21} and SLS~\cite{b42} introduce shape or scale-location constraints to improve localization. SCTransNet~\cite{b64} further enhances feature interactions through spatial-channel cross-transformer modeling.  Dim2Clear~\cite{b90} exploits spatial-frequency feature enhancement, while EAA-Net~\cite{b73} addresses aliasing in infrared target representations. More recently, SAIST~\cite{b68} and Text-IRSTD~\cite{b110} incorporate language guidance, InvDet~\cite{b109} explores reconstruction-guided feature preservation, and NS-FPN~\cite{b101} suppresses background interference. Despite these advances, most existing approaches primarily focus on enhancing target representations or suppressing noisy features. However, stronger feature responses do not necessarily indicate more reliable target predictions, particularly when background structures exhibit target-like characteristics.

In complex infrared scenes, cloud edges, thermal hotspots, and structured background clutter can produce target-like responses that resemble genuine small targets~\cite{b59,b94,b105,b113}. Although background masking and feature purification mitigate such interference~\cite{b112,b101}, salient clutter responses may still persist across multiple scales~\cite{b64}. As illustrated in Fig.~\ref{fig_big}, genuine targets and background interference may exhibit similarly strong raw responses, but differ in their verified evidence. Hence, response magnitude alone cannot establish target reliability, motivating explicit evidence verification.

To this end, we propose VBE-UNet, a Target Evidence Chain Verification Network for infrared small target detection. Rather than directly treating enhanced feature responses as reliable target cues, VBE-UNet assesses candidate reliability from three complementary perspectives: local background deviation, cross-scale consistency, and perturbation-induced prediction sensitivity. Specifically, the Counterfactual Background Explanation (CBE) module approximates a target-absent background reference using ring-shaped neighborhood statistics and extracts background-relative residuals as local target evidence. The joint evidence verification mechanism evaluates the consistency of candidate responses across adjacent decoder scales and estimates their sensitivity to controlled feature perturbations through changes in auxiliary predictions. These complementary cues are integrated to assess evidence reliability rather than independently establish target identity. Finally, the Evidence-Guided Recomposition (EGR) module selectively refines decoder features according to the joint evidence, preserving target-relevant responses while suppressing unsupported background activations.

Our main contributions are summarized as follows:
\begin{itemize}
    \item We propose a local background explanation mechanism that models target evidence through ring-shaped neighborhood statistics and normalized differential features.

    \item We develop a joint evidence verification mechanism that integrates cross-scale evidence support and perturbation-based contribution verification to guide reliable feature recomposition.

    \item Extensive experiments on IRSTD-1K, NUDT-SIRST, and NUAA-SIRST demonstrate the effectiveness of VBE-UNet in improving detection accuracy and suppressing false alarms.
\end{itemize}

\begin{figure*}[t!]
    \centering
    \includegraphics[width=0.95\textwidth]{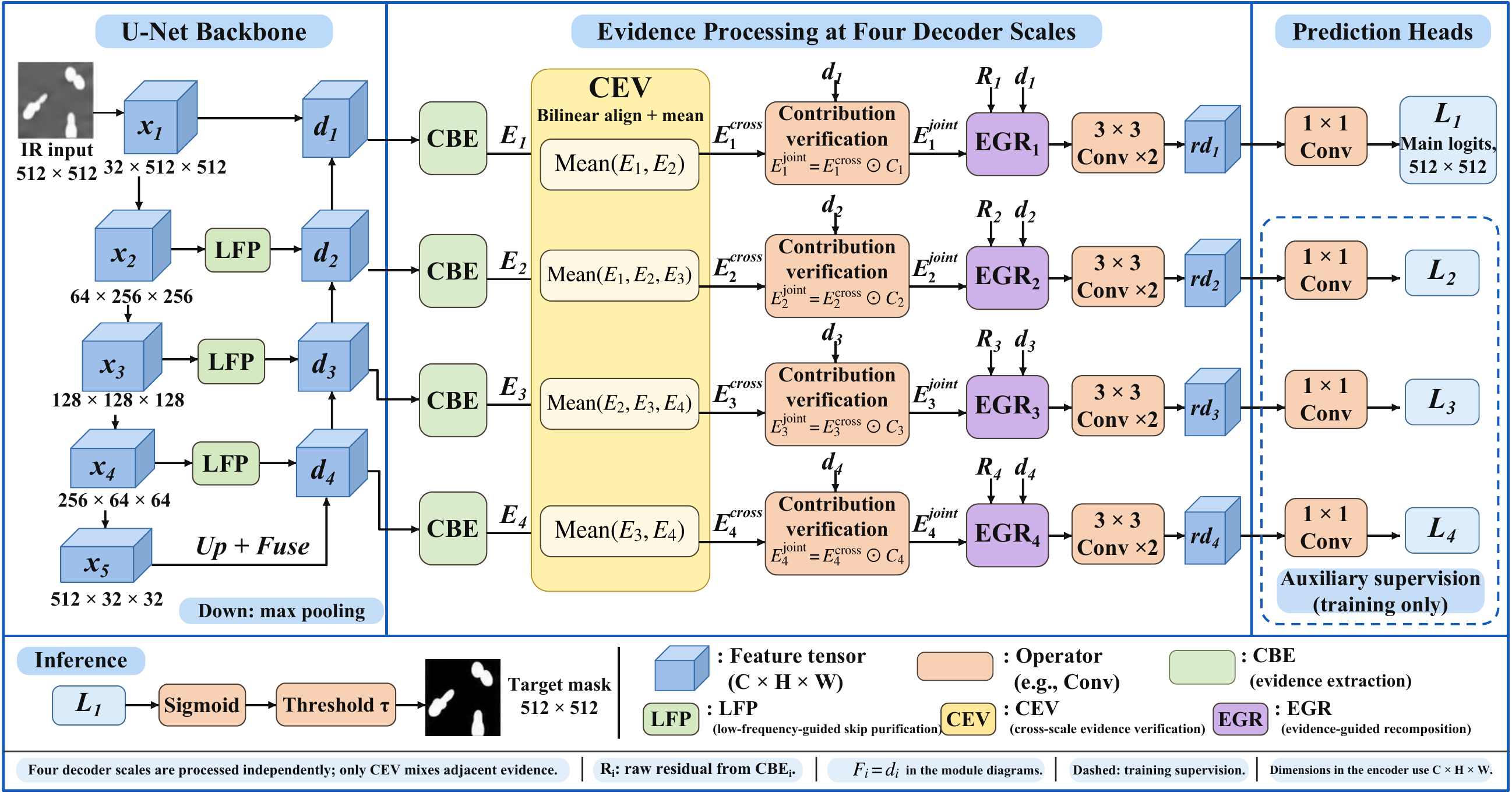}
    \caption{Overall architecture of VBE-UNet. A U-Net backbone with low-frequency-guided skip purification extracts multi-scale features. The proposed framework models local background evidence, integrates cross-scale support and contribution verification, and guides residual feature recomposition for infrared small target segmentation.}
    \label{fig_overall}
\end{figure*}

\section{Related Work}
\label{sec:related}

\subsection{Model-driven Infrared Small Target Detection}

Traditional infrared small target detection methods typically exploit handcrafted assumptions about target contrast, spatial structure, and background statistics. Background filtering methods use morphological operators or local statistics to suppress slowly varying regions~\cite{b52}, while local contrast measures highlight intensity differences between candidate targets and their neighborhoods~\cite{b53,b103,b104}. Another representative line formulates targets as sparse outliers against a low-rank background. The infrared patch-image (IPI) model~\cite{b20} performs patch-based low-rank and sparse decomposition, and subsequent tensor-based approaches incorporate additional spatial and nonlocal correlations~\cite{b55,b56}. RPCANet~\cite{b95} bridges model-driven priors and deep learning by unfolding robust PCA into trainable stages. Global contrast modeling further uses a decomposition-derived baseline to detect targets that may not be locally salient~\cite{b113}. Although these approaches are interpretable and often effective without extensive training data, their handcrafted assumptions can be unreliable in scenes containing strong edges, heterogeneous clutter, and rapidly varying background structures~\cite{b105}.


\subsection{Deep Representation Learning for Infrared Small Target Detection}

Deep learning-based methods learn target-sensitive representations through hierarchical feature extraction, multi-scale fusion, and structural supervision. Compared with handcrafted methods, these networks can integrate wider contextual information and learn nonlinear target-background distinctions from data. ACM~\cite{b30} uses asymmetric contextual modulation to integrate high-level semantics with low-level details, while DNANet~\cite{b38} and UIU-Net~\cite{b58} employ nested feature interactions to mitigate the loss of weak target signals. AGPCNet~\cite{b94} models multi-scale context, while CMNet~\cite{b63} uses cross-layer guidance to improve target representations. ISNet~\cite{b21} incorporates shape-aware guidance to improve target localization, whereas SLS~\cite{b42} introduces scale- and location-sensitive learning. 
SCTransNet~\cite{b64} strengthens spatial-channel feature interactions, while SAIST~\cite{b68} and Text-IRSTD~\cite{b110} exploit language guidance. InvDet~\cite{b109} instead uses reconstruction guidance to preserve target information. Diffusion-based continuous feature modeling~\cite{b60} further addresses the representation of dim, low-contrast targets. In addition to accuracy-oriented architectures, IRPruneDet~\cite{b89} uses wavelet-regularized channel pruning to improve computational efficiency while retaining important target information. Despite substantial progress, these methods predominantly emphasize representation quality, feature interaction, or model efficiency. However, increased representation capacity may also preserve responses from challenging background distractors, while the reliability of individual candidate responses is not always examined explicitly.


\subsection{Background Interference Suppression and Target Evidence Reliability}

Suppressing background interference is essential for avoiding false alarms, especially when clutter exhibits target-like intensity and structure~\cite{b57,b105}. Such interference is not limited to random noise, as cloud boundaries and thermal hotspots may also exhibit coherent spatial patterns that are difficult to suppress without affecting weak targets. IRSAM~\cite{b87} adapts the Segment Anything Model to infrared imagery by introducing diffusion-based feature processing for noise suppression and structural preservation, together with multi-granularity decoding. PBT~\cite{b59} incorporates background-aware features. BGM~\cite{b112} masks background regions during training, whereas LRRNet~\cite{b111} learns low-rank background representations through compression and reconstruction. NS-FPN~\cite{b101} suppresses noise propagation via low-frequency-guided purification and spiral-aware sampling. These approaches improve background modeling or suppress interfering features, but neither strong responses nor agreement across scales necessarily establishes that a candidate corresponds to a genuine target. In particular, structured clutter may produce stable activations that survive enhancement and denoising. Unlike conventional feature enhancement or noise suppression, VBE-UNet combines background-relative residual modeling, cross-scale evidence support, and perturbation-based contribution verification to guide feature recomposition.

\section{Method}
\label{sec:method}

\subsection{Overall Architecture}
As illustrated in Fig.~\ref{fig_overall}, VBE-UNet adopts a U-Net encoder--decoder. Infrared inputs are normalized with ImageNet statistics. A U-Net backbone with low-frequency-guided purification~\cite{b101} on its three deeper skip connections produces four decoder features $\{F_i\}_{i=1}^4$, indexed from high to low resolution. Counterfactual Background Explanation (CBE) estimates local evidence and residuals; Joint Evidence Verification combines cross-scale consistency with probe-response sensitivity; and Evidence-Guided Recomposition (EGR) then refines features according to the verified evidence. Thus, evidence verification precedes evidence-conditioned enhancement, while LFP serves as preliminary feature purification.

\begin{figure}[t!]
    \centering
    \includegraphics[width=\columnwidth]{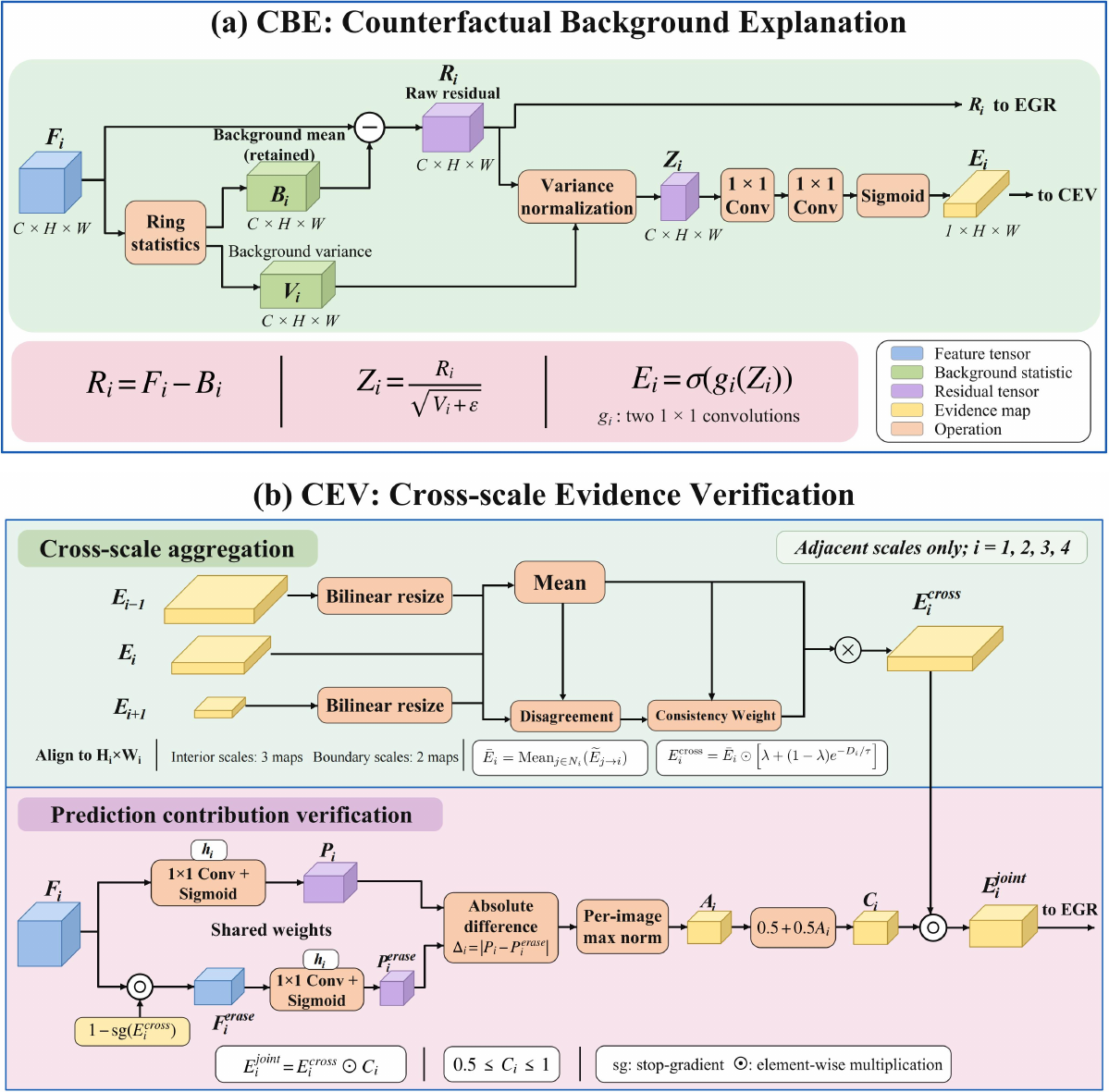}
    \caption{Target evidence modeling and verification. (a) Counterfactual Background Explanation (CBE) estimates target evidence through background-relative residual modeling. (b) Cross-scale Evidence Verification (CEV) integrates cross-scale evidence aggregation and prediction contribution verification to produce jointly verified evidence.}
    \label{fig_verification}
\end{figure}

\subsection{Local Background Evidence Modeling}
CBE models each candidate relative to an annular background reference, providing a counterfactual proxy rather than a guaranteed target-absent feature state. Overlapping receptive fields may introduce some target information into this reference.

For $F_i\in\mathbb{R}^{c_i\times H_i\times W_i}$, let $\mathcal{R}(x)=\mathcal{N}_{15}(x)\setminus\mathcal{N}_{5}(x)$ denote the annular neighborhood at $x$, with $\mathcal{N}_r$ an $r\times r$ window. Only valid positions $\mathcal{R}_i^v(x)=\mathcal{R}(x)\cap\Omega_i$ are used to avoid padding bias, and $m_i(x)=|\mathcal{R}_i^v(x)|$. The channel-wise background statistics are computed as

\begin{equation}
\begin{aligned}
B_i(x)&=\frac{1}{m_i(x)}
\sum_{u\in\mathcal{R}_i^{v}(x)}F_i(u),\\
V_i(x)&=\max\!\left(
\frac{1}{m_i(x)}
\sum_{u\in\mathcal{R}_i^{v}(x)}F_i^2(u)
-B_i^2(x),\epsilon\right).
\end{aligned}
\end{equation}
The background-relative residual and its normalized counterpart are
\begin{equation}
R_i=F_i-B_i,\qquad Z_i=\frac{R_i}{\sqrt{V_i+\epsilon}},
\end{equation}
where $\epsilon$ ensures numerical stability. $R_i$ preserves signed background deviations, while $Z_i$ normalizes them according to
local background variation. 

A lightweight projection converts the normalized representation into a spatial evidence map:
\begin{equation}
E_i=\sigma\!\left(g_i(Z_i)\right),
\end{equation}
where $g_i$ comprises two $1\times1$ convolutions with intermediate batch normalization and ReLU, and $\sigma$ is sigmoid. Thus, $E_i$ encodes learned background-relative evidence rather than a direct target probability. We retain $R_i$ separately for subsequent recomposition, avoiding premature enhancement of potentially ambiguous responses.

\begin{figure}[t!]
    \centering
    \includegraphics[width=\columnwidth]{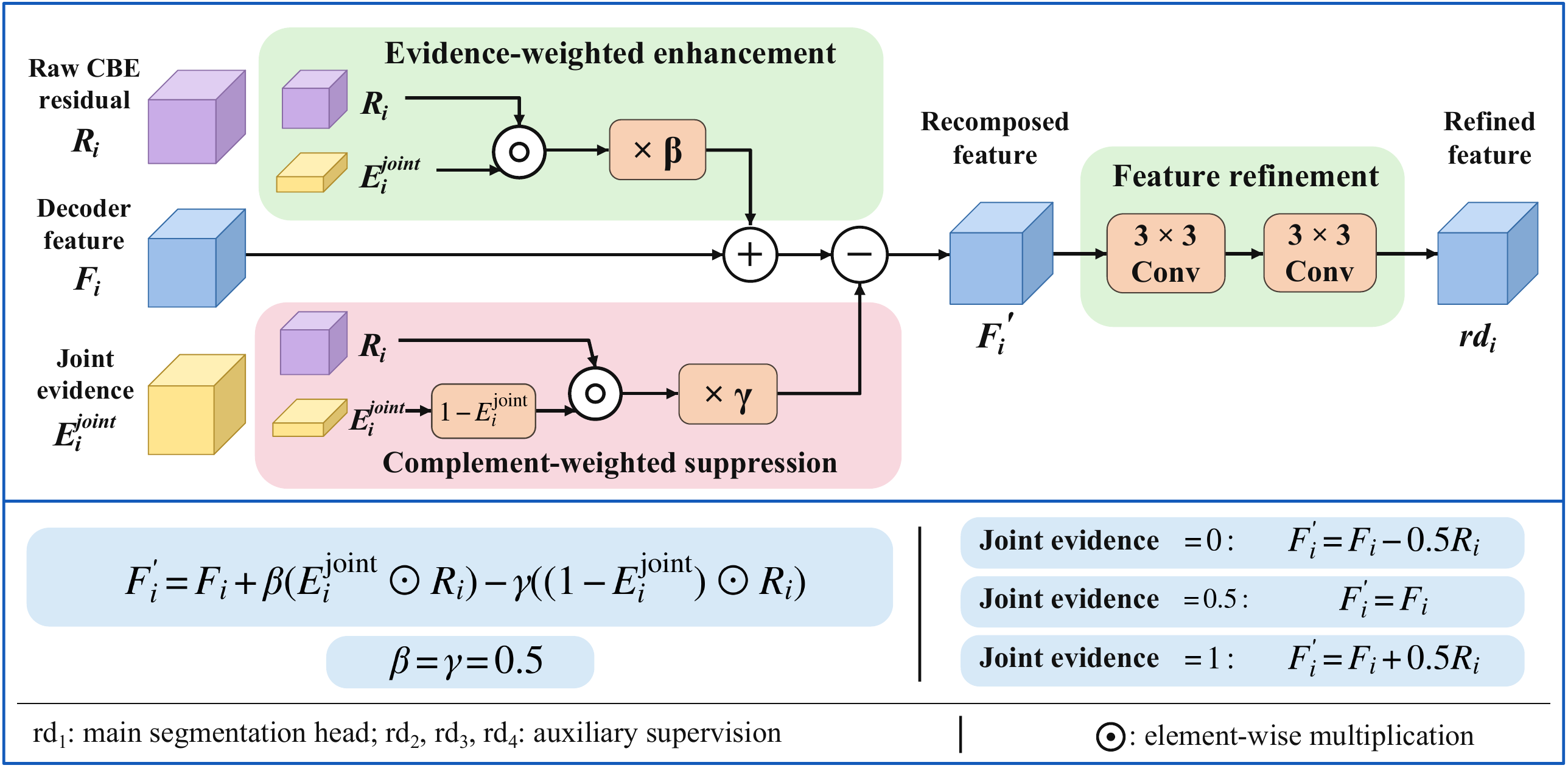}
    \caption{Evidence-guided recomposition (EGR). Joint evidence $E_i^{\mathrm{joint}}$ controls complementary enhancement and suppression of background-relative residuals $R_i$. The recomposed feature $F_i'$ is subsequently refined by two $3\times3$ convolutions to produce $rd_i$.}
    \label{fig_egr}
\end{figure}

\subsection{Joint Evidence Verification}
Although CBE captures background-relative deviations, strong residual responses may also arise from structured clutter. We therefore further assess candidate reliability from two complementary perspectives: cross-scale consistency and perturbation-induced prediction sensitivity, as illustrated in Fig.~\ref{fig_verification}(b). The former evaluates whether local evidence receives consistent support across adjacent decoder scales, while the latter measures changes in auxiliary predictions under evidence-guided feature attenuation. Together, these cues provide a reliability estimate for subsequent feature refinement rather than independently establishing target identity.

\noindent\textbf{Cross-scale evidence verification.}
Evidence maps from adjacent decoder scales are first aligned and aggregated to obtain cross-scale support:
\begin{equation}
\bar E_i(x)= \frac{1}{|N_i|} \sum_{j\in N_i} \widetilde E_{j\rightarrow i}(x),
\end{equation}
where $N_i=\{i-1,i,i+1\}\cap\{1,2,3,4\}$ and $\widetilde E_{j\rightarrow i}=\operatorname{Align}_{j\rightarrow i}(E_j)$
denotes bilinear alignment.

To explicitly assess cross-scale consistency, we measure the disagreement among aligned evidence maps:
\begin{equation}
D_i(x)=\frac{1}{|N_i|} \sum_{j\in N_i} \left( \widetilde E_{j\rightarrow i}(x)-\bar E_i(x) \right)^2.
\end{equation}
The consistency-weighted evidence is then computed as
\begin{equation}
E_i^{\mathrm{cross}}(x)= \bar E_i(x) \left[ \lambda+(1-\lambda) \exp\left(-\frac{D_i(x)}{\tau}\right) \right],
\end{equation}
where $\tau>0$ controls sensitivity to disagreement and $\lambda\in[0,1]$ preserves a fraction of the aggregated evidence. 
This consistency-aware weighting suppresses responses with conflicting cross-scale evidence while retaining mutually supported activations. However, structured clutter may also exhibit cross-scale consistency. We therefore further assess how the auxiliary prediction responds to evidence-guided feature attenuation.

\noindent\textbf{Perturbation-based Contribution Verification.}
Cross-scale consistency alone cannot establish target reliability, as structured clutter may also produce coherent responses. We therefore introduce contribution verification to assess the predictive influence of candidate features through evidence-guided attenuation:
\begin{equation}
F_i^{\mathrm{erase}}=
F_i\odot\big(1-\operatorname{sg}(E_i^{\mathrm{cross}})\big),
\end{equation}
where $\odot$ denotes element-wise multiplication and $\operatorname{sg}$ stops gradients through the attenuation mask. The same auxiliary prediction probe $h_i$ is applied before and after attenuation:
\begin{equation}
P_i=\sigma(h_i(F_i)),\quad
P_i^{\mathrm{erase}}=
\sigma(h_i(F_i^{\mathrm{erase}})).
\end{equation}
We quantify the perturbation-induced prediction change as
\begin{equation}
A_i(x)=
\frac{|P_i(x)-P_i^{\mathrm{erase}}(x)|}
{\max\left(\max_u |P_i(u)-P_i^{\mathrm{erase}}(u)|,
\epsilon\right)}.
\end{equation}
Here, $A_i$ provides a perturbation-based estimate of feature contribution by measuring changes in the auxiliary prediction under evidence-guided attenuation. This estimate reflects relative predictive influence rather than signed or causal contribution to the final segmentation output. Since the attenuation is conditioned on cross-scale evidence, the contribution estimate is interpreted jointly with cross-scale support rather than as independent evidence of target correctness. We incorporate it into the cross-scale evidence through
\begin{equation}
C_i=0.5+0.5A_i,\qquad
E_i^{\mathrm{joint}}=
E_i^{\mathrm{cross}}\odot C_i.
\end{equation}
The bounded gate $C_i\in[0.5,1]$ adaptively discounts cross-scale evidence according to the estimated predictive contribution. The resulting joint evidence is used to guide subsequent feature recomposition.

\subsection{Evidence-Guided Recomposition}
Evidence estimation alone does not determine how a candidate should affect the decoder representation. EGR translates the joint evidence into a controlled update of the background-relative residual, rather than applying uniform enhancement to the entire feature. As shown in Fig.~\ref{fig_egr}, we formulate
\begin{equation}
F_i'=F_i+\beta(E_i^{\mathrm{joint}}\odot R_i)
-\gamma((1-E_i^{\mathrm{joint}})\odot R_i),
\label{eq:egr}
\end{equation}
where $\beta$ and $\gamma$ control residual enhancement and suppression, respectively. We set $\beta=\gamma=0.5$, yielding
$F_i'=F_i+(E_i^{\mathrm{joint}}-0.5)\odot R_i$. This enables evidence-conditioned refinement while preserving the original decoder feature.

Each recomposed feature is refined by two $3\times3$ convolutions and projected to segmentation logits:
\begin{equation}
rd_i=\operatorname{Refine}_i(F_i'),\qquad
L_i=\operatorname{Conv}_{1\times1}(rd_i).
\end{equation}
The main prediction is $\widehat{Y}=\sigma(L_1)$, while $L_2$, $L_3$, and $L_4$ supply auxiliary supervision. This design ties residual adjustment to the jointly assessed evidence at each decoding level.

\begin{figure*}[t!]
    \centering
    \includegraphics[width=\textwidth]{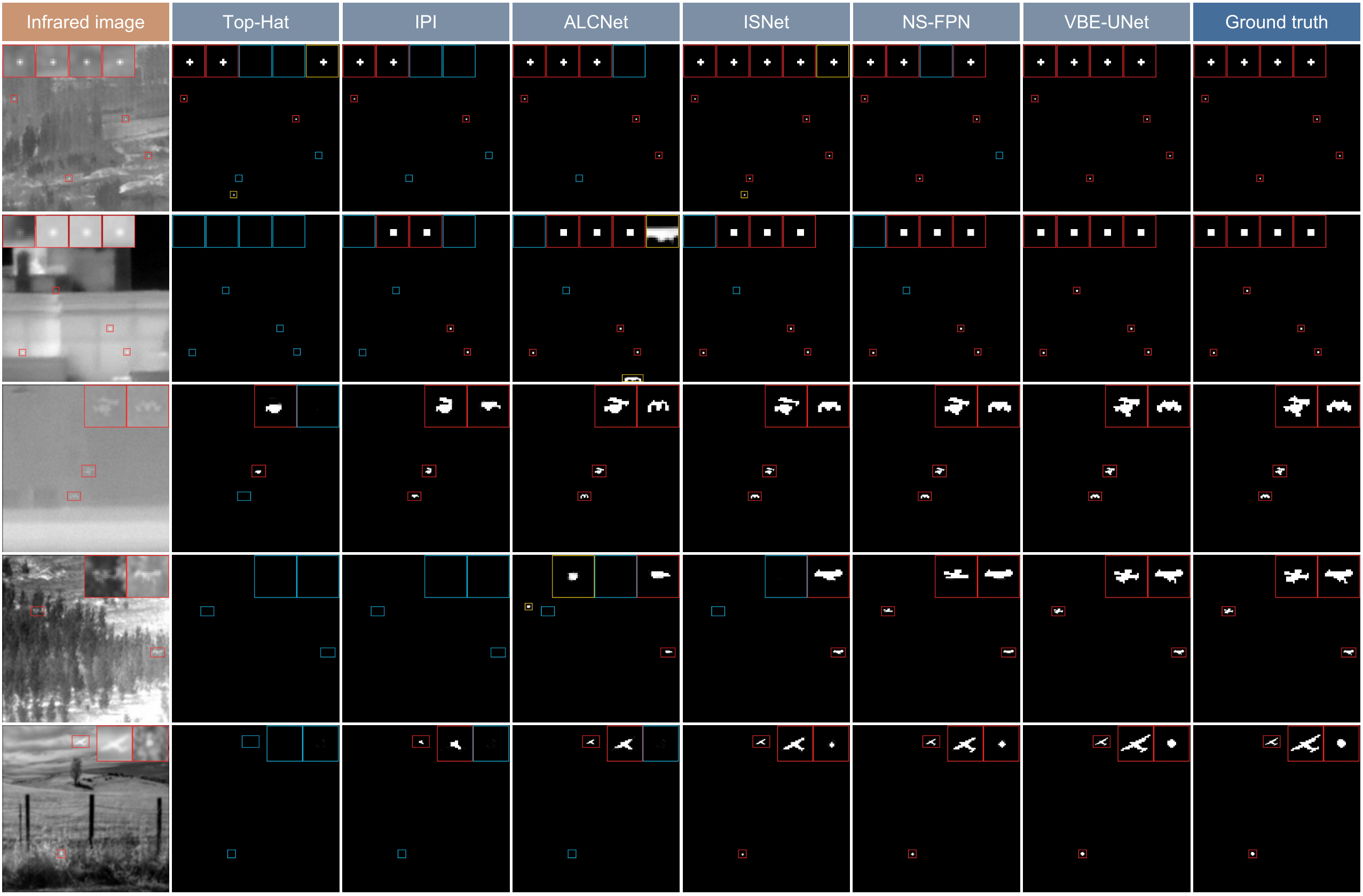}
    \caption{Qualitative comparison of VBE-UNet with representative infrared small target detection methods. Red, blue, and yellow boxes indicate correctly detected targets, missed targets, and false alarms, respectively. Enlarged target regions are shown at the top of each example.}
    \label{fig5}
\end{figure*}

\subsection{Training Objective}
We optimize segmentation and evidence learning jointly. Let $\mathcal{L}_{\mathrm{BD}}=\mathcal{L}_{\mathrm{WBCE}}+\mathcal{L}_{\mathrm{Dice}}$. The primary segmentation loss is
\begin{equation}
\mathcal{L}_{\mathrm{seg}}=
\mathcal{L}_{\mathrm{SLS}}(L_1,Y)+
0.5\mathcal{L}_{\mathrm{BD}}(L_1,Y),
\end{equation}
where $\mathcal{L}_{\mathrm{SLS}}$ follows scale-and-location-sensitive supervision~\cite{b42}, and the positive-class weight in weighted BCE is 10. Its scale and location components are gradually activated over ten epochs after an initial five-epoch soft-IoU phase. Three native-resolution auxiliary logits are supervised with
\begin{equation}
\mathcal{L}_{\mathrm{aux}}=\sum_{i=2}^{4}w_i\big[
\mathcal{L}_{\mathrm{SLS}}(L_i,Y_i)+
0.25\mathcal{L}_{\mathrm{BD}}(L_i,Y_i)\big],
\end{equation}
where $(w_2,w_3,w_4)=(0.30,0.20,0.10)$, and $Y_i$ is obtained by adaptive max-pooling the ground-truth mask to the corresponding logit resolution to preserve small targets.

For evidence supervision, we resize the ground-truth mask by nearest-neighbor interpolation and the detached main prediction probability by bilinear interpolation to each evidence resolution. Foreground pixels form $\Omega_i^+$; hard negatives $\Omega_i^-$ are background pixels whose detached prediction scores exceed 0.35, or, if none exceed this threshold, up to 1024 highest-scoring background pixels. The evidence penalty is
\begin{equation}
\mathcal{L}_{\mathrm{ev}}(M_i)=
\frac{1}{|\Omega_i^+|}\sum_{x\in\Omega_i^+}(1-M_i(x))+
\frac{1}{|\Omega_i^-|}\sum_{x\in\Omega_i^-}M_i(x),
\end{equation}
where terms over empty sets are omitted. The loss is first averaged over valid images and then across valid decoder levels. The full objective is
\begin{equation}
\begin{aligned}
\mathcal{L}_{\mathrm{total}}={}&\mathcal{L}_{\mathrm{seg}}+\mathcal{L}_{\mathrm{aux}}\\
&+0.15\overline{\mathcal{L}}_{\mathrm{ev}}(E^{\mathrm{joint}})
+0.03\overline{\mathcal{L}}_{\mathrm{ev}}(E)\\
&+0.03\overline{\mathcal{L}}_{\mathrm{ev}}(C),
\end{aligned}
\end{equation}
where the overbar denotes the aforementioned multi-level averaging. 
The sensitivity loss supervises the bounded sensitivity gate $C_i=0.5+0.5A_i$, which represents a probe-based contribution estimate
rather than a calibrated target probability. For samples containing both positive and selected negative pixels, the unweighted evidence penalty satisfies $\mathcal{L}_{\mathrm{ev}}(C_i)=0.5\mathcal{L}_{\mathrm{ev}}(A_i)+0.5$, so its constant offset does not affect gradients.

\section{Experiments}
\subsection{Datasets}

We evaluate our method on three widely used infrared small target detection datasets: NUAA-SIRST~\cite{b30}, NUDT-SIRST~\cite{b38}, and IRSTD-1K~\cite{b21}. NUAA-SIRST contains 427 real infrared images, which are divided into 213 training images and 214 testing images. NUDT-SIRST consists of 1,327 synthetic infrared images with diverse target scales and background conditions, with 663 images for training and 664 images for testing. IRSTD-1K contains 1,001 real infrared images with complex backgrounds, of which 800 images are used for training and 201 images for testing. We follow the commonly adopted dataset splits for fair comparison with existing methods.

\begin{table*}[t]
    \centering
    \footnotesize
    \setlength{\tabcolsep}{3pt}

    \caption{Quantitative comparison of VBE-UNet with state-of-the-art methods on IRSTD-1K, NUDT-SIRST, and NUAA-SIRST. Results are reported in terms of nIoU (\%), mIoU (\%), $P_d$ (\%), and $F_a$ ($10^{-6}$).}
    \label{tab:comparison}

    \begin{tabular*}{\textwidth}{
        @{\extracolsep{\fill}}
        l *{4}{c}
        @{\hspace{6pt}}
        *{4}{c}
        @{\hspace{6pt}}
        *{4}{c}
        @{}
    }

        \toprule
        Method
        & \multicolumn{4}{c}{IRSTD-1K\cite{b21}}
        & \multicolumn{4}{c}{NUDT-SIRST\cite{b38}}
        & \multicolumn{4}{c}{NUAA-SIRST\cite{b30}} \\

        \cmidrule(lr){2-5}
        \cmidrule(lr){6-9}
        \cmidrule(lr){10-13}

        & \textit{nIoU}$\uparrow$
        & \textit{mIoU}$\uparrow$
        & \textit{Pd}$\uparrow$
        & \textit{Fa}$\downarrow$
        & \textit{nIoU}$\uparrow$
        & \textit{mIoU}$\uparrow$
        & \textit{Pd}$\uparrow$
        & \textit{Fa}$\downarrow$
        & \textit{nIoU}$\uparrow$
        & \textit{mIoU}$\uparrow$
        & \textit{Pd}$\uparrow$
        & \textit{Fa}$\downarrow$ \\

        \midrule

        ALCNet\cite{b12}
        & 62.68 & 63.21 & 89.25 & 27.71
        & 72.89 & 75.42 & 91.19 & 30.40
        & 73.12 & 74.31 & 97.34 & 20.21 \\

        IRPruneDet~\cite{b89}
        & 62.71 & 64.54 & 91.74 & 16.04
        & 74.12 & 78.31 & 92.18 & 29.42
        & 74.30 & 75.12 & 98.61 & 2.96 \\

        ISNet\cite{b21}
        & 63.88 & 65.31 & 92.59 & 27.92
        & 77.86 & 79.81 & 92.59 & 24.65
        & 77.29 & 80.02 & 99.02 & 4.61 \\

        FC3Net~\cite{b88}
        & 63.59 & 64.98 & 92.93 & 15.73
        & 78.23 & 80.25 & 94.86 & 16.48
        & 72.46 & 74.22 & 99.12 & 6.56 \\

        DNANet\cite{b38}
        & 65.71 & 68.02 & 91.84 & 17.61
        & 79.98 & 80.65 & 96.93 & 12.78
        & 77.05 & 79.26 & 98.48 & 5.35 \\

        ABCNet\cite{b22}
        & 67.09 & 68.46 & 92.99 & 17.32
        & 80.34 & 80.82 & 97.01 & 12.51
        & 79.32 & 81.01 & 98.00 & 9.51 \\

        SCTransNet\cite{b64}
        & 68.15 & 68.53 & 93.03 & 16.56
        & 81.08 & 82.57 & 97.50 & 12.09
        & 76.18 & 77.50 & 96.95 & 13.92 \\

        MSHNet\cite{b42}
        & 67.16 & 69.32 & 93.88 & 15.03
        & 80.55 & 82.88 & 97.99 & 11.77
        & 76.89 & 80.48 & 98.52 & 10.43 \\

        ACANet\cite{b43}
        & 67.82 & 70.13 & 94.03 & 14.72
        & 80.59 & 83.24 & 98.21 & 11.32
        & 77.29 & 80.73 & 99.28 & 8.46 \\

        Light-SGMTLM\cite{b40}
        & 67.95 & 70.58 & 94.15 & 14.52
        & 80.15 & 83.52 & 98.22 & 11.15
        & 70.61 & 71.03 & 94.29 & 15.84 \\

        SAIST\cite{b68}
        & 68.15 & 71.25 & 94.28 & 14.35
        & 80.42 & 83.67 & 98.25 & 11.03
        & 78.74 & 80.82 & 99.56 & 0.87 \\

        IRSAM~\cite{b87}
        & 68.21 & 70.85 & 94.39 & 14.27
        & 80.29 & 83.35 & 98.64 & 11.29
        & 78.32 & 79.86 & 98.34 & 1.74 \\
        NS-FPN~\cite{b101}
        & 69.29 & 71.38 & 94.83 & 10.58
        & 80.83 & 84.15 & 98.87 & 10.61
        & 78.75 & 80.26 & 100 & 1.60 \\
        \textbf{VBE-UNet} & \textbf{69.86} & \textbf{72.75} & \textbf{95.74} & \textbf{9.94} & \textbf{81.42} & \textbf{84.36} & \textbf{99.47} & \textbf{9.82} & \textbf{80.03} & \textbf{81.54} & \textbf{100} & \textbf{0.79} \\
        \bottomrule

    \end{tabular*}
\end{table*}

\subsection{Implementation Details}
VBE-UNet is trained in PyTorch on a single RTX 5090 GPU for 500 epochs with a batch size of 4 and $512\times512$ inputs. We use AdamW with an initial learning rate of $10^{-4}$, weight decay of $0.01$, polynomial decay, and a five-epoch warm-up.

We set $\beta=\gamma=\lambda=0.5$ and $\tau=0.05$, with CBE window sizes of 5 and 15. Each scale uses a two-layer $1\times1$ convolutional probe with weights shared between original and attenuated features. LFP~\cite{b101} is applied to three deeper skip connections. We hold out 10\% of training data for validation and use a fixed inference threshold of 0.5 without test-time augmentation.

\begin{figure}[t!]
    \centering
    \includegraphics[width=\columnwidth]{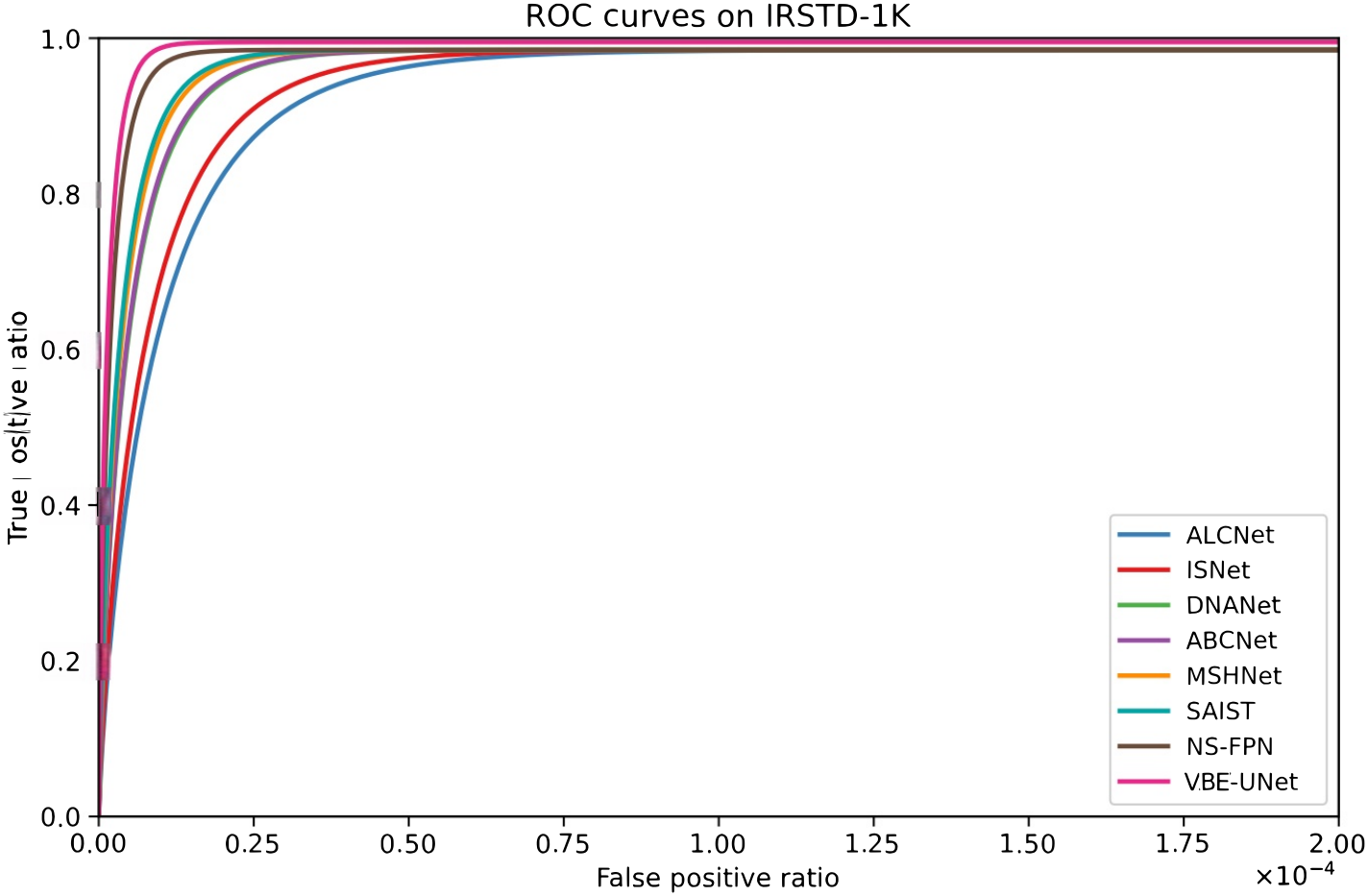}
    \caption{ROC curve comparison of different methods
    on the IRSTD-1K dataset.}
    \label{fig_roc}
\end{figure}

\subsection{Evaluation Metrics}
We adopt four standard metrics: mIoU, nIoU, detection probability ($P_d$), and false-alarm rate ($F_a$). The pixel-level metrics are
defined as
\begin{equation}
\begin{aligned}
\mathrm{mIoU}
&=\frac{\sum_i TP_i}
{\sum_i(TP_i+FP_i+FN_i)},\\
\mathrm{nIoU}
&=\frac{1}{N}\sum_i
\frac{TP_i}{TP_i+FP_i+FN_i}.
\end{aligned}
\end{equation}
Here, $N$ denotes the number of test images. $P_d$ measures the fraction of detected targets, while $F_a$ measures the proportion of false-positive pixels.

\subsection{Quantitative Comparison}
We compare VBE-UNet with 13 state-of-the-art methods on three IRSTD benchmarks. As shown in Table~\ref{tab:comparison}, VBE-UNet achieves
the highest nIoU and mIoU across all three datasets. On IRSTD-1K, it achieves 69.86\% nIoU and 72.75\% mIoU, surpassing the strongest competing results by 0.57 and 1.37 percentage points, respectively. On NUDT-SIRST, our method achieves the best performance across all four metrics. On NUAA-SIRST, it obtains the highest nIoU and mIoU and the lowest false-alarm rate, while matching the best detection probability of 100\%. These results demonstrate the effectiveness of VBE-UNet in target segmentation and false-alarm suppression.

We further compare the receiver operating characteristic (ROC) curves on IRSTD-1K, as shown in Fig.~\ref{fig_roc}. VBE-UNet consistently achieves a higher true positive rate (TPR) at low false positive rates (FPR) than competing methods, demonstrating its effectiveness in target
detection and false-alarm suppression.

\begin{table}[t]
    \centering
    \caption{Ablation study on IRSTD-1K.}
    \label{tab:ablation}
    \footnotesize
    \setlength{\tabcolsep}{3pt}
    \begin{tabular*}{\columnwidth}{
        @{\extracolsep{\fill}}lcccc@{}}
        \toprule
        Variants & nIoU$\uparrow$ & mIoU$\uparrow$
        & $P_d\uparrow$ & $F_a\downarrow$ \\
        \midrule

        Baseline
        & 62.34 & 65.18 & 88.26 & 21.47 \\

        w/o CBE
        & 66.72 & 69.43 & 92.15 & 14.68 \\

        w/o Joint Verification
        & 68.41 & 71.26 & 94.32 & 12.37 \\

        w/o EGR
        & 67.58 & 70.64 & 93.57 & 13.25 \\
        
        w/o Cross-scale
        & 66.03 & 67.42 & 91.85 & 13.27 \\

        w/o Contribution Verification
        & 67.13 & 69.28 & 92.93 & 11.73 \\
        w/o LFP & 69.14 & 72.28 & 94.93 & 10.31 \\
        \textbf{Full VBE-UNet}
        & \textbf{69.86} & \textbf{72.75}
        & \textbf{95.74} & \textbf{9.94} \\

        \bottomrule
    \end{tabular*}
\end{table}

\subsection{Qualitative Comparison}
Fig.~\ref{fig5} presents qualitative comparisons with representative methods under challenging infrared scenes. Traditional methods suffer from missed detections and background interference, while several deep learning-based methods produce incomplete target predictions or false alarms. In contrast, VBE-UNet achieves more accurate target localization and shape recovery while suppressing spurious background responses, demonstrating its effectiveness in complex scenes.

\begin{table}[t]
    \centering
    \caption{Evidence reliability comparison.}
    \label{tab:evidence_reliability}
    \small
    \begin{tabular*}{\columnwidth}{@{\extracolsep{\fill}}lcc@{}}
        \toprule
        Evidence Representation & AUROC $\uparrow$ & AUPRC $\uparrow$ \\
        \midrule
        Local Evidence (CBE)    & 0.90 & 0.45 \\
        Joint Verified Evidence & \textbf{0.94} & \textbf{0.53} \\
        \bottomrule
    \end{tabular*}
\end{table}

\subsection{Ablation Study}

\noindent\textbf{Ablation Settings.}
All variants share the same backbone, training schedule, and segmentation losses. The baseline excludes CBE, Joint Verification, and EGR. Without LFP, the original skip features are directly forwarded to the decoder. Without CBE, evidence is predicted directly from decoder features, with $E_i=\sigma(g_i(F_i))$ and $R_i=F_i$. Without Joint Verification, we set $E_i^{\mathrm{joint}}=E_i$. Without EGR, we set $F_i'=F_i$. Removing cross-scale aggregation replaces $E_i^{\mathrm{cross}}$ with $E_i$, while removing contribution verification sets $C_i=1$. Loss terms associated with removed branches are omitted.

\noindent\textbf{Component Ablation.}
As shown in Table~\ref{tab:ablation}, the full VBE-UNet improves nIoU from 62.34\% to 69.86\% over the baseline. Removing CBE, Joint Verification, or EGR reduces nIoU to 66.72\%, 68.41\%, and 67.58\%, respectively, demonstrating the contribution of each component. 
Removing LFP slightly decreases nIoU from 69.86\% to 69.14\%, indicating that the proposed framework remains effective without LFP.

\noindent\textbf{Verification Ablation.}
We further evaluate the two mechanisms within Joint Evidence Verification. Removing cross-scale aggregation or probe-response sensitivity decreases nIoU from 69.86\% to 66.03\% and 67.13\%, respectively, while increasing the false alarm rate. These results demonstrate the complementary benefits of cross-scale evidence support and probe-response sensitivity. Cross-scale verification contributes more to nIoU improvement, while contribution verification provides complementary gains.

\begin{figure}[t]
    \centering
    \includegraphics[width=\columnwidth]{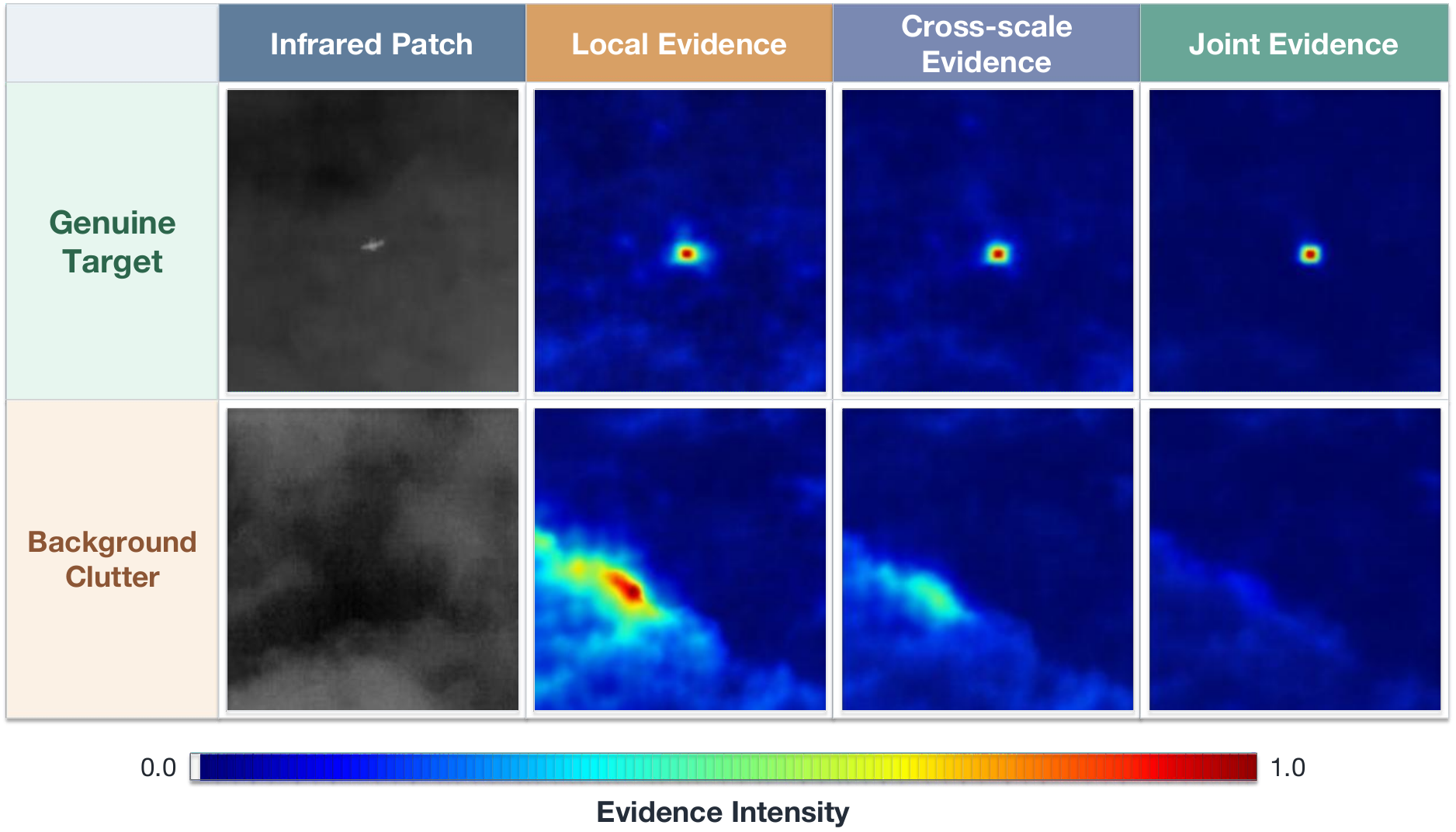}
    \caption{Visualization of target evidence at different verification stages. Local, cross-scale, and jointly verified evidence are compared for genuine infrared targets and target-like background clutter.}
    \label{fig:evidence_visualization}
\end{figure}

\noindent\textbf{Evidence Discriminability.}To evaluate evidence discriminability, we compute pixel-level AUROC and AUPRC for local and jointly verified evidence against ground-truth masks on IRSTD-1K. As shown in Table~\ref{tab:evidence_reliability}, joint verification improves AUROC from 0.90 to 0.94 and AUPRC from 0.45 to 0.53.

\noindent\textbf{Evidence Visualization.}
Fig.~\ref{fig:evidence_visualization} visualizes the evidence responses at different verification stages. Compared with local evidence, jointly verified evidence preserves concentrated responses around genuine targets while suppressing spurious activations from background clutter. These results further demonstrate the effectiveness of joint evidence verification in distinguishing reliable target evidence from background interference.

\begin{table}[t]
    \centering
    \caption{Computational complexity and inference time on $256\times256$ inputs (batch size = 1).}
    \label{tab:complexity256}
    \footnotesize
    \setlength{\tabcolsep}{4pt}
    \resizebox{\columnwidth}{!}{%
        \begin{tabular}{@{}lccccc@{}}
            \toprule
            Metric & UIUNet~\cite{b58} & ABCNet~\cite{b22} & AGPCNet~\cite{b94} & MSHNet~\cite{b42} & VBE-UNet \\
            \midrule
            Params (M) & 50.54 & 106.99 & 12.43 & 16.57 & \textbf{9.42} \\
            FLOPs (G) & 217.84 & 166.27 & 863.32 & 891.15 & \textbf{44.71} \\
            Time (ms) & 123.750 & 55.562 & 108.125 & 216.250 & \textbf{40.128} \\
            \bottomrule
        \end{tabular}%
    }
\end{table}

\subsection{Computational Complexity}

We compare the computational complexity and inference efficiency of VBE-UNet with representative methods using $256\times256$ inputs. As shown in Table~\ref{tab:complexity256}, VBE-UNet achieves the lowest parameter count (9.42M) and FLOPs (44.71G) among the compared methods. It also achieves the fastest inference time of 40.128 ms, outperforming the other methods in computational efficiency. These results demonstrate that the proposed evidence verification framework maintains relatively low computational overhead while achieving competitive detection performance.

\section{Conclusion}
In this paper, we propose VBE-UNet, an evidence-based reliability assessment framework for infrared small target detection. Rather than relying solely on enhanced response magnitude, VBE-UNet jointly assesses local background deviation, cross-scale consistency, and perturbation-induced prediction sensitivity to guide feature refinement. Specifically, counterfactual background explanation models local target evidence, joint evidence verification integrates cross-scale support and probe-response sensitivity, and evidence-guided recomposition adaptively refines decoder features to improve target-background discrimination. Extensive experiments on three public benchmarks demonstrate the effectiveness of VBE-UNet in improving detection accuracy and reducing false alarms. Ablation studies and evidence reliability analysis further validate the contributions of the proposed mechanisms.

\section*{LLM Usage Statement}

OpenAI's ChatGPT was used as an assistive tool to refine
author-produced content, including manuscript writing and
editing, related work and citations, code and implementation,
experimental descriptions and analysis, figures and visual
materials, data-related procedures, and the presentation of
research ideas and conclusions. All AI-assisted suggestions
were reviewed and verified by the authors. The authors retain
full responsibility for the research methodology, experimental
results, scientific claims, and all submitted content.
{
    \small
    \bibliographystyle{ieeenat_fullname}
    \bibliography{main}
}

\end{document}